\documentclass[10pt,twocolumn,letterpaper]{article}

\usepackage{cvpr}              

\usepackage{graphicx}
\usepackage{amsmath}
\usepackage{amssymb}
\usepackage{booktabs}

\usepackage[pagebackref,breaklinks,colorlinks]{hyperref}

\usepackage[capitalize]{cleveref}
\crefname{section}{Sec.}{Secs.}
\Crefname{section}{Section}{Sections}
\Crefname{table}{Table}{Tables}
\crefname{table}{Tab.}{Tabs.}

\usepackage[pagebackref,breaklinks,colorlinks]{hyperref}
\usepackage{algorithm}
\usepackage{algorithmic}

\usepackage{multirow}
\usepackage{placeins}
\usepackage{color, colortbl}
\definecolor{mygray}{gray}{0.926}
\usepackage[accsupp]{axessibility}

\def\cvprPaperID{3428}
\def\confName{CVPR}
\def\confYear{2023}

\begin{document}

\title{Regularized Vector Quantization for Tokenized Image Synthesis}

\author{Jiahui Zhang\textsuperscript{\rm 1} \quad
Fangneng Zhan\textsuperscript{\rm 2} \quad
Christian Theobalt\textsuperscript{\rm 2} \quad
Shijian Lu\thanks{Corresponding author, E-mail: shijian.lu@ntu.edu.sg}\ \textsuperscript{\rm 1} \\
\textsuperscript{\rm 1} Nanyang Technological University
\quad
\textsuperscript{\rm 2} Max Planck Institute for Informatics
}

\maketitle

\begin{abstract}
Quantizing images into discrete representations has been a fundamental problem in unified generative modeling. Predominant approaches learn the discrete representation either in a deterministic manner by selecting the best-matching token or in a stochastic manner by sampling from a predicted distribution. However, deterministic quantization suffers from severe codebook collapse and misalignment with inference stage while stochastic quantization suffers from low codebook utilization and perturbed reconstruction objective. This paper presents a regularized vector quantization framework that allows to mitigate above issues effectively by applying regularization from two perspectives. The first is a prior distribution regularization which measures the discrepancy between a prior token distribution and the predicted token distribution to avoid codebook collapse and low codebook utilization. The second is a stochastic mask regularization that introduces stochasticity during quantization to strike a good balance between inference stage misalignment and unperturbed reconstruction objective. In addition, we design a probabilistic contrastive loss which serves as a calibrated metric to further mitigate the perturbed reconstruction objective. Extensive experiments show that the proposed quantization framework outperforms prevailing vector quantization methods consistently across different generative models including auto-regressive models and diffusion models.
\end{abstract}

\section{Introduction}
With the prevalence of multi-modal image synthesis \cite{baltruvsaitis2018multimodal,ramesh2021zero,zhan2021multimodal,yu2022towards} and Transformers \cite{vaswani2017attention},
unifying data modeling regardless of data modalities has attracted increasing interest from the research communities.
Aiming for a generic data representation across different data modalities, discrete representation learning \cite{oord2017neural,rolfe2016discrete} plays a significant role in the unified modeling.
In particular, vector quantization models (e.g., VQ-VAE \cite{oord2017neural} and VQ-GAN \cite{esser2020taming}) emerge as a promising family for learning generic image representations by discretizing images into discrete tokens.
With the tokenized representation,
generative models such as auto-regressive model \cite{gregor2014deep,esser2020taming} and diffusion model \cite{ho2020denoising,dhariwal2021diffusion} can be applied to accommodate the dependency of the sequential tokens for image generation, which is referred as \textit{tokenized image synthesis} under this context.

\begin{figure}[t]
\centering
\includegraphics[width=1.0\linewidth]{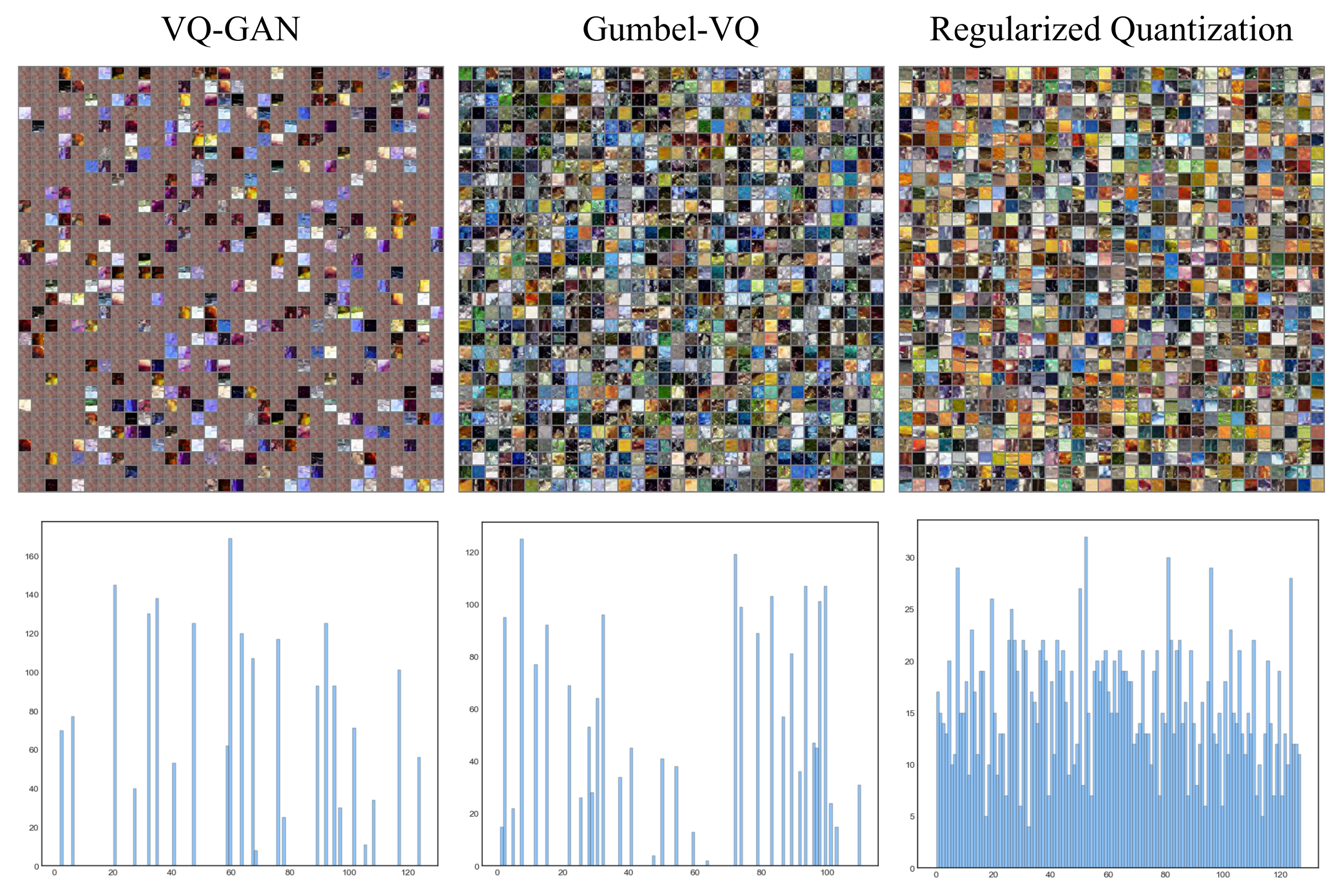}
\caption{
Visualization of codebook (first row) and illustration of codebook utilization (second row) on ADE20K dataset \cite{zhou2017ade20k}.
VQ-GAN \cite{esser2020taming} severely suffers from codebook collapse as most codebook embeddings are invalid values.
Gumbel-VQ \cite{baevski2019vq} learns valid values for all codebook embeddings,
while only a small number of embeddings are actually used for quantization as illustrated in codebook utilization.
As a comparison, the proposed regularized quantization prevents codebook collapse and achieves full codebook utilization.
The codebook visualization method is provided in the supplementary file.
}
\label{im_codebook}
\end{figure}

Vector quantization models can be broadly grouped into \textbf{deterministic} quantization and \textbf{stochastic} quantization according to the selection of discrete tokens.
Specifically, typical deterministic methods like VQ-GAN \cite{esser2020taming} directly select the best-matching token via Argmin or Argmax, while stochastic methods like Gumbel-VQ \cite{baevski2019vq} select a token by stochastically sampling from a predicted token distribution.
On the other hand, deterministic quantization suffers from codebook collapse \cite{roy2018theory}, a well-known problem where large portion of codebook embeddings are invalid with near-zero values as shown in Fig. \ref{im_codebook} (first row).
In addition, deterministic quantization is misaligned with the inference stage of generative modeling, where the tokens are usually randomly sampled instead of selecting the best matching one.
Instead, stochastic quantization samples tokens according to a predicted token distribution with Gumbel-Softmax \cite{jang2016categorical,baevski2019vq}, which allows to avoid codebook collapse and mitigate inference misalignment.
However, although most codebook embeddings are valid values in stochastic quantization, only a small part is actually utilized for vector quantization as shown in Fig. \ref{im_codebook} (second row), which is dubbed as low codebook utilization.
Besides, as stochastic methods randomly sample tokens from a distribution, the image reconstructed from the sampled tokens is usually not well aligned with the original image, leading to perturbed reconstruction objective and unauthentic image reconstruction.

In this work, we introduce a regularized quantization framework that allows to prevent above problems effectively via regularization from two perspectives.
Specifically,
to avoid codebook collapse and low codebook utilization where only a small number of codebook embeddings are valid or used for quantization, 
we introduce a \textbf{prior distribution regularization} by assuming a uniform distribution as the prior for token distribution.
As the posterior token distribution can be approximated by the quantization results, we can measure the discrepancy between the prior token distribution and posterior token distribution.
By minimizing the discrepancy during training, the quantization process is regularized to use all the codebook embeddings, which prevents the predicted token distribution from collapse into a small number of codebook embeddings.

As deterministic quantization suffers from inference stage misalignment and stochastic quantization suffers from perturbed reconstruction objective, we introduce a \textbf{stochastic mask regularization} to strike a good balance between them.
Specifically, the stochastic mask regularization randomly masks certain ratio of regions for stochastic quantization, while leaving the unmasked regions for deterministic quantization.
This introduces uncertainty for the selection of tokens and results of quantization, which narrows the gap with the inference stage of generative modelling where tokens are selected randomly.
We also conduct thorough and comprehensive experiments to analyze the selection of masking ratio for optimal image reconstruction and generation.

On the other hand, with the randomly sampled tokens, the stochastically quantized region will suffer from perturbed reconstruction objective.
The perturbed reconstruction objective mainly results from the target for perfect reconstruction of the original image from randomly sampled tokens.
Instead of naively enforcing a perfect image reconstruction with L1 loss,
we introduce a contrastive loss for elastic image reconstruction, which mitigates the perturbed reconstruction objective significantly.
Similar to PatchNCE \cite{park2020contrastive,zhan2022modulated}, the contrastive loss treats the patch at the same spatial location as positive pairs and others as negative pairs.
By pushing the positive pairs closer and pulling negative pairs away, the elastic image reconstruction can be achieved.
Another issue with the randomly sampled tokens is that they tend to introduce perturbation of different scales in the reconstruction objective,
We thus introduce a Probabilistic Contrastive Loss (PCL) that adjusts the pulling force of different regions according to the discrepancy between the sampled token embedding and the best-matching token embedding.

The contributions of this work can be summarized in three aspects. 
First, we present a regularized quantization framework that introduces a prior distribution regularization to prevent codebook collapse and low codebook utilization.
Second, we propose a stochastic mask regularization which mitigates the misalignment with the inference stage of generative modelling.
Third, we design a probabilistic contrastive loss that achieves elastic image reconstruction and mitigates the perturbed objective adaptively for different regions with stochastic quantization.

\section{Related Work}

\subsection{Vector Quantization}

As introduced in Oord \emph{et al.} \cite{oord2017neural},
vector quantization aims to represent the data with entries of a learnt codebook (i.e., tokens), which achieves discrete and compressed representation.
According to the selection mechanism of discrete tokens, 
the quantization methods can be grouped into deterministic quantization and stochastic quantization.

\textbf{Deterministic Quantization.}
With a predicted token distribution or probability,
deterministic quantization aims to select the best-matching token through Argmin or Argmax.
Typically, VQ-VAE \cite{oord2017neural} proposes to quantize the encoded feature into discrete token by looking up the nearest neighbour (i.e., Argmin) entry in a learned codebook.
As the operation of Argmax is not differentiable, there is no real gradient defined for the encoder. VQ-VAE adopts Straight Through Estimation (STE) \cite{bengio2013estimating} by copying the gradient from the decoder to the encoder.
During training, the encoder, decoder and codebook are optimized driven by a reconstruction loss and a codebook embedding loss \cite{oord2017neural}.
Following this line of deterministic quantization, strenuous effort has been made to improve the quantization performance, including Exponential Moving Averages (EMA) \cite{oord2017neural,razavi2019generating} for stable updating of codebook, multi-scale hierarchical organization \cite{razavi2019generating} for higher synthetic coherence and fidelity, adversarial loss and perceptual loss \cite{esser2020taming} for improved perceptual quality, integrated quantization \cite{zhan2022auto} for condition generation, translation-equivariant quantization with orthogonal codebook embeddings \cite{shin2021translation}, Transformer structure for quantization \cite{yu2021vector}.

\textbf{Stochastic Quantization.}
Instead of naively selecting the best-matching token, stochastic quantization aims to sample token from a predicted token distribution. 
As the sampling operation is not differentiable, certain reparameterization trick (e.g., Gumbel-Softmax \cite{jang2016categorical}) should be applied for gradient backpropagation. 
For instance, 
VQ-Wave2Vec \cite{baevski2019vq} introduces stochastic quantization with Gumbel-Softmax reparameterization trick to learn the discrete representation of audios.
Similarly, DALL-E \cite{ramesh2021zero} leverages Gumbel-Softmax to represent images with tokens, followed by a Transformer to auto-regressively model the dependency between tokens for diverse image synthesis.
Inspired by the connection between Exponential Moving Averages (EMA) \cite{oord2017neural} and Expectation Maximization (EM) algorithm \cite{moon1996expectation},
Roy \emph{et al.} \cite{roy2018theory} introduce soft EM algorithm for quantization by performing Monte-Carlo Expectation Maximization \cite{wei1990monte} on the probability distribution of discrete latent variables.
The sampling from a Gumbel-Softmax distribution exactly approximates the sampling from the token distribution as proved in \cite{maddison2014sampling}.

\subsection{Tokenized Image Synthesis}

With a tokenized representation of images, generative models can be applied to the discrete tokens for image synthesis \cite{esser2020taming,zhan2022auto,yu2021diverse}.
For example, PixelRNN and PixelCNN~\cite{van2016pixel} employ LSTM~\cite{hochreiter1997long} and masked convolutions to model inter-dependencies of pixels autoregressively.
By learning a compressed tokenized representation,
VQ-VAE~\cite{oord2017neural} achieves compelling generation quality with PixelCNN.
Recently, Transformer \cite{vaswani2017attention} emerges as a powerful paradigm for sequence modeling.
Chen~\emph{et al.}~\cite{chen2020generative} leverage Transformer to model the sequence dependency of image pixels.
DALL-E \cite{ramesh2021zero} designs a discrete VAE for learning tokenized representation and models the image tokens with a Transformer to achieve text-to-image generation.
Esser~\emph{et al.}~\cite{esser2020taming} propose a VQ-GAN to learn a rich discrete representation and utilize the Transformer to efficiently model token distributions for high-resolution images synthesis.

In addition to auto-regressive models, diffusion models \cite{sohl2015deep,ho2020denoising,dhariwal2021diffusion} can also be applied to model discrete tokens.
For instance, D3PMs~\cite{austin2021structured} adopts discrete diffusion to estimate the density of image pixels and achieves low-resolution image synthesis.
With the learned discrete tokens in VQ-VAE, VQ-Diffusion \cite{gu2021vector} achieves compelling text-to-image generation performance via a discrete diffusion process.

\begin{figure*}[t]
\centering
\includegraphics[width=1.0\linewidth]{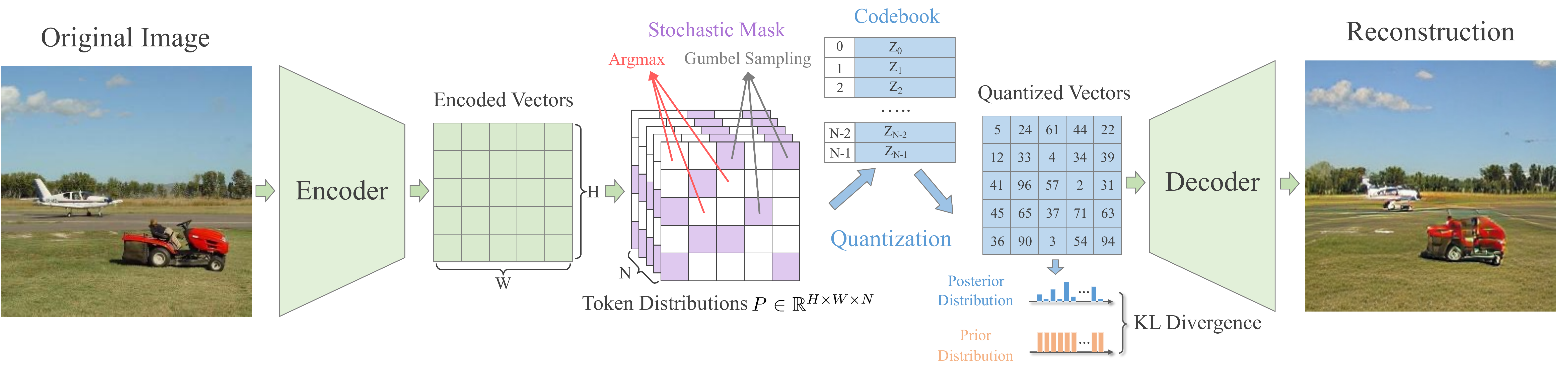}
\caption{
The framework of the proposed regularized quantization. A stochastic mask (indicated by purple regions) is applied to the \textit{Predicted Token Distributions} to specify the region for stochastic sampling. Then the \textit{Encoded Vectors} can be represented by the selected codebook embeddings, which produces the \textit{Quantized Vectors} for image \textit{Reconstruction}. A KL divergence is measured between the posterior token distribution and the prior token distribution to avoid codebook collapse and low codebook utilization.
}
\label{im_mask}
\end{figure*}

\section{Method}

As illustrated in Fig. \ref{im_mask}, our regularized quantization framework combines deterministic quantization and stochastic quantization, which consists of an encoder $E$, a decoder $G$, and a codebook $\mathcal{Z} = \{z_n\}_{n=1}^N \in \mathbb{R}^{N \times d}$, where $N$ is the size of the codebook and $d$ is the dimension of embeddings.
Given an input image $X$, the encoder is employed to produce a spatial collection of token distributions $x_{i}\in \mathbb{R}^N, i\in[1,H\times W]$, where $H\times W$ is the size of spatial vectors.
Then, each encoded vector is mapped into a discrete token according to the predicted token distribution, which yields tokenized representation (i.e., indices of codebook embeddings).
The codebook embeddings associated with the indices are finally fed into the decoder to reconstruct the input image.

With a trained vector quantization framework, we can represent images in terms of the codebook indices (i.e., tokens).
With the discrete tokens of images, generative models such as auto-regressive model \cite{gregor2014deep} and diffusion model \cite{ho2020denoising} can be applied to build the dependency between tokens.
At inference stage of generative modeling, a sequence of tokens can be sampled for image synthesis.
By mapping the sequence of tokens back to their corresponding codebook embeddings, an image can be readily generated by feeding the embeddings into the decoder.

\begin{figure}[t]
\centering
\includegraphics[width=1.0\linewidth]{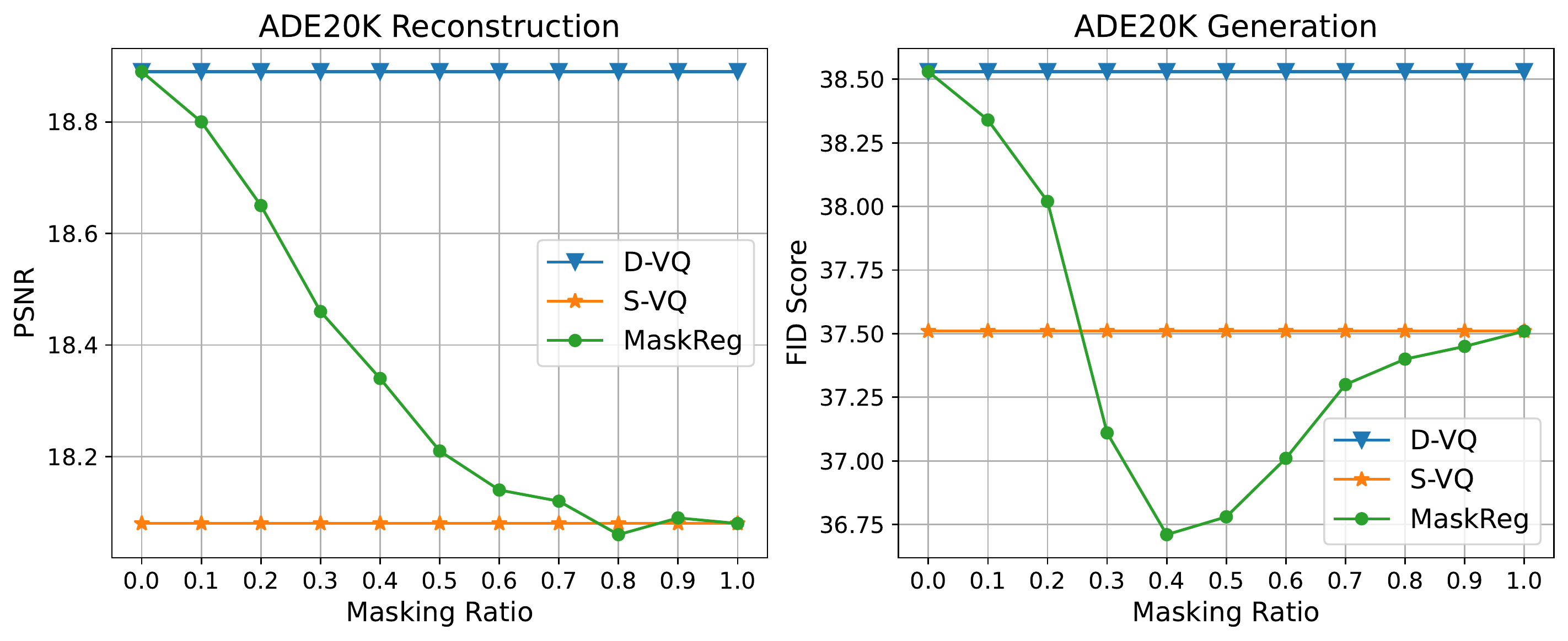}
\caption{
The effect of different masking ratios on image reconstruction and image generation on ADE20K dataset.
VQ-GANs with different quantization methods are used for image tokenization and auto-regressive model is used for image synthesis.
D-VQ, S-VQ, and MaskReg denote deterministic vector quantization, stochastic vector quantization, and the proposed stochastic mask regularization, respectively.
The quality of image reconstruction is evaluated by PSNR, the quality of image generation is evaluated by FID.
Note good PSNR score doesn't indicate good generation quality.
}
\label{im_graph}
\end{figure}

\subsection{Prior Distribution Regularization}

Prevailing vector quantization models usually severely suffer from codebook collapse or low codebook utilization, where only a small number of codebook embeddings are valid or used for quantization.
We thus propose a prior distribution regularization to regularize the vector quantization process.
Specifically, we assume a prior distribution for the tokens used for quantization.
Ideally, the prior distribution is expected to be a uniform discrete distribution denoted by $P_{prior}=[1/N, 1/N, \cdots, 1/N], P_{prior}\in \mathbb{R}^{N}$, which means
all codebook embeddings can be used uniformly and their corresponding information capacity can be maximized according to the principle of maximum entropy.
During quantization, as image features of size $H\times W$ are mapped to corresponding tokens, the predicted quantization result of each feature can be represented by a one-hot vector $p_i, i\in [1, H*W]$.
Thus, the posterior token distribution $P_{post}$ can be approximated by the average of all one-hot vector as $P_{post}=\sum_{i=1}^{H\times W} p_i / (H\times W)=[p_1, p_2, \cdots, p_N]$.
Then, the discrepancy between the prior token distribution and predicted token distribution can be measured by KL divergence as below:
\begin{equation}
    \mathcal{L}_{kl} = KL(P_{post}, P_{prior}) = -\sum_{n}^{N} p_n \log \frac{1/N}{p_n}
\end{equation}
By minimizing the KL divergence $\mathcal{L}_{kl}$, the vector quantization can be effectively regularized to avoid codebook collapse and low codebook utilization.

\subsection{Stochastic Mask Regularization}

On the other hand, the deterministic quantization which selects the most probable tokens will lead to misalignment with the inference stage of generative modelling, where tokens are sampled randomly according to the predicted distribution.
Instead, stochastic quantization enables to introduce stochasticity during quantization which helps to mitigate the misalignment with inference stage.
Nevertheless, stochastic quantization tends to incur perturbed reconstruction objective as the sampled tokens may not match the original image.
As a result, the generation quality (FID) of stochastic quantization presents marginal improvement compared with deterministic quantization as shown in Fig. \ref{im_graph}.  
To strike a good balance between unperturbed reconstruction objective and inference stage misalignment,
we design a stochastic mask regularization which combines the deterministic quantization and stochastic quantization by applying a stochastic mask to coordinate the image regions for stochastic quantization and deterministic quantization as shown in Fig. \ref{im_mask}.

Specifically, with the predicted token probability $P\in \mathbb{R}^{H\times W \times N}$ for all encoded vectors, we randomly set a mask $M \in \mathbb{R}^{H\times W}$ with `1' to indicate the regions for token sampling with Gumbel-softmax and `0' to indicate the regions for selecting the best-matching tokens with Argmax.
Denoting the vectors quantized via Argmax and Gumbel sampling as $X_{argmax}$ and $X_{gumbel} \in \mathbb{R}^{H\times W \times N}$, respectively, the reconstruction objective can be formulated as below:
\begin{align*}
\mathcal{L}_{rec} = \| X - G(X_{argmax}*(1-M) + X_{gumbel}*M) \|_1 ,
\end{align*}
where $G$ denotes the decoder.
As shown in Fig. \ref{im_graph}, 
comprehensive experiments are conducted to analyze the effect with different masking ratios, and a masking ratio of 40\% is proved to yield the best image reconstruction \& generation quality (i.e., best FID). 
The effect of different masking ratios on other datasets is provided in the supplementary material.
As both Argmax and Gumbel sampling operations are non-differentiable, we apply reparameterization trick by replacing Argmax operation with Softmax and Gumbel with Gumbel-Softmax in gradient back-propagation.
The pseudo code of the forward \& backward propagation of the proposed regularized quantization is given in Algorithm \ref{code}.

\begin{algorithm}[ht]
  \caption{Pseudo-code of forward \& backward propagation in vector quantization with the proposed stochastic mask regularization.}
    \begin{algorithmic}
      \footnotesize
      \STATE \textbf{Deterministic Quantization Region.}
      \vspace{1pt}
      \STATE \textbf{Input:} encoded vector $x_{ij}$, token distribution $P_{ij}=[p_1, p_2, \cdots, p_N]$, codebook $\mathcal{Z}$.
      
        \STATE \quad \textbf{Forward propagation:}
        \STATE \quad 1. index = Argmax($P_{ij}$)
        \STATE \quad 2. index\_hard = One\_Hot(index)
        \STATE \quad 3. quantized $\hat{x}_{ij}$ = Matmul(index\_hard, $\mathcal{Z}$)
    
        \STATE \quad \textbf{Backward propagation:}
        \STATE \quad 1. index\_soft = Softmax($P_{ij}$)
        \STATE \quad 2. quantized ${\hat{x}_{ij}}$ = Matmul(index\_soft, $\mathcal{Z}$)
      \STATE \textbf{Output:} quantized ${\hat{x}_{ij}}$.
    \end{algorithmic}
    
    \vspace{5pt}
    
    \begin{algorithmic}
    \footnotesize
      \STATE \textbf{Stochastic Quantization Region.}
      \vspace{1pt}
      \STATE \textbf{Input:} encoded vector $x_{ij}$, token distribution $P_{ij}=[p_1, p_2, \cdots, p_N]$, codebook $\mathcal{Z}$, gumbels $\sim$ Gumbel(0, 1).
      
        \STATE \quad \textbf{Forward propagation:}
        \STATE \quad 1. index = Argmax($P_{ij}$+gumbels)
        \STATE \quad 2. index\_hard = One\_Hot(index)
        \STATE \quad 3. quantized $\hat{x}_{ij}$ = Matmul(index\_hard, $\mathcal{Z}$)
        \STATE \quad \textbf{Backward propagation:}
        \STATE \quad 1. index\_soft = Softmax($P_{ij}$+gumbels)
        \STATE \quad 2. quantized $\hat{x}_{ij}$ = Matmul(index\_soft, $\mathcal{Z}$)
      \STATE \textbf{Output:} quantized $\hat{x}_{ij}$.
    \end{algorithmic}
    
  \label{code}
\end{algorithm}

\begin{figure}[t]
\centering
\includegraphics[width=1.0\linewidth]{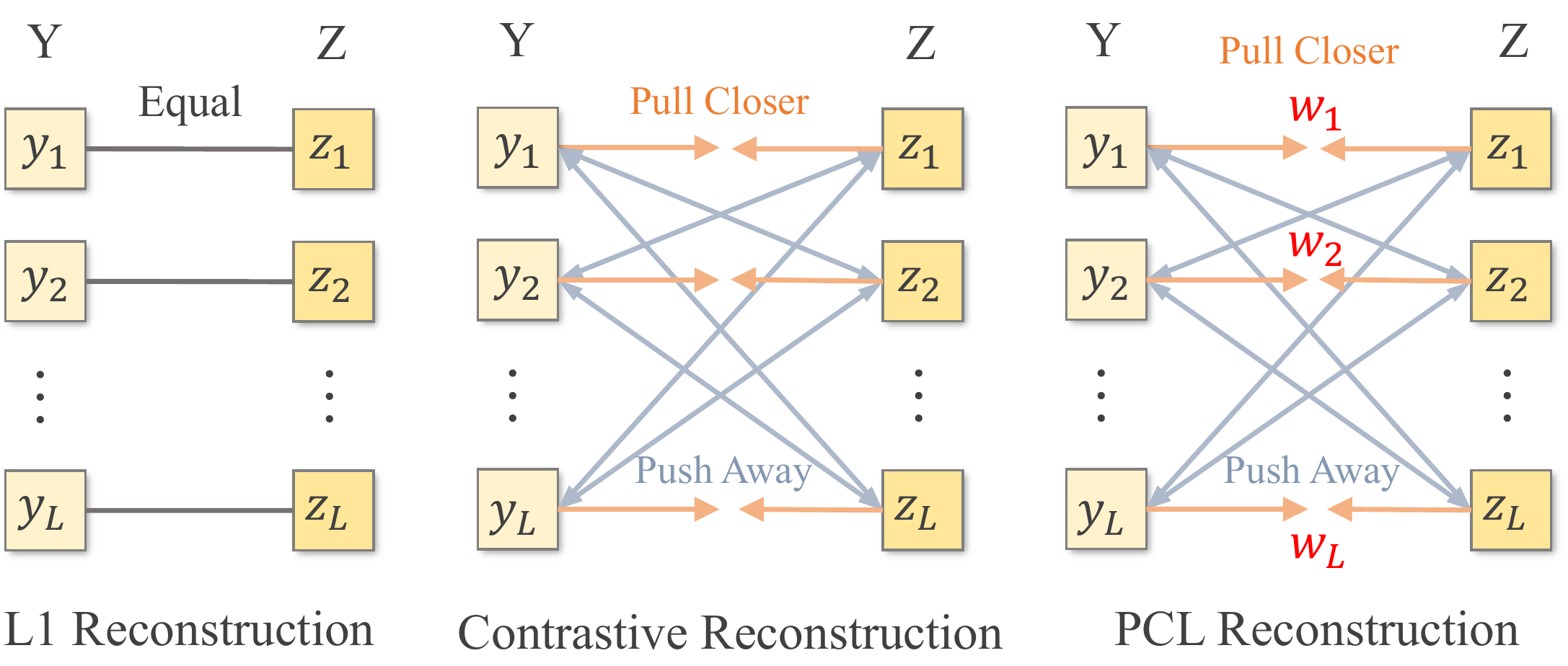}
\caption{
Comparisons of L1 loss, vanilla contrastive loss, and our proposed probabilistic contrastive loss (PCL) for image reconstruction.  PCL introduces adaptive weights $\{w_i\}_{i=1}^{N}$ to positive pairs for better representation learning and elastic image reconstruction.
}
\label{im_pcl}
\end{figure}

\subsection{Probabilistic Contrastive Loss}

The stochastic mask regularization mitigates the misalignment with the inference stage.
However, for the image region with stochastic quantization, the model training still suffers from perturbed reconstruction objective caused by the randomly sampled tokens.
Thus, we propose a probabilistic contrastive loss (PCL) to mitigate the perturbed objective in the region with stochastic quantization.
As the perturbed objective results from the target for perfect reconstruction of original images with L1 loss, the proposed PCL achieves \textit{elastic image reconstruction}~\footnote{'Elastic' means relative / contrastive approximation of the original image instead of perfect reconstruction.} in the stochastic quantization region through contrastive learning. 

Instead of forcing perfect image reconstruction, contrastive learning aims to maximize the mutual information between corresponding images by pulling selected positive pairs closer and pushing negative pairs away as shown in Fig. \ref{im_pcl}.
Following the Noise Contrastive Estimation framework \cite{oord2018representation} in PatchNCE \cite{park2020contrastive,zhan2022marginal}, image features in the same spatial location of the original and reconstructed image are regarded as positive pairs and others are negative pairs in PCL.
Thus, vanilla contrastive loss $\mathcal{L}_{cl}$ for image reconstruction can be formulated as:
\begin{equation}
\label{patchnce}
\mathcal{L}_{cl} = -  \frac{1}{L} \sum_{i=1}^L \log
\frac{e^{y_i \cdot z_i / \tau}} 
{e^{y_i \cdot z_i / \tau} + \sum_{\substack{j=1 \\ j\neq i}}^L e^{ y_i \cdot z_j / \tau}},
\end{equation}
where $Y = [y_1, y_2, \cdots, y_L]$ and $Z = [z_1, z_2, \cdots, z_L]$ are extracted feature patches from the original image and reconstructed image respectively, $\tau$ is the temperature parameter, $L$ is the number of image features.
As proved in \cite{esser2020taming,dong2021peco}, perceptual loss \cite{johnson2016perceptual} helps to keep good perceptual quality for image reconstruction.
We thus employ pre-trained VGG-19 network \cite{simonyan2014very} to extract multi-layer image features ($relu1\_2, relu2\_2, relu3\_3, relu4\_3, relu5\_3$) from original and reconstructed images to construct contrastive learning pairs. Note, the contrastive loss is used alongside the perceptual loss during training.

\textbf{Probabilistic Contrast.}
With the stochasticity in Gumbel sampling, the sampled tokens tend to present varying discrepancies with the best-matching one as selected by Argmax.
Intuitively, a sampled token with larger discrepancy with the best-matching one will yield severer objective perturbation.
Thus, the pulling force between the original and reconstructed images should be adaptive with respect to the perturbation magnitude for optimal contrastive learning.
We introduce the Probabilistic Contrastive Loss (PCL) which employs \textbf{weighting parameters} $\{w_i\}_{i=1}^{L}$ to adjust the pulling force of different features according to the token sampling results (i.e., perturbation magnitude) as shown in Fig. \ref{im_pcl}.
The weighting parameter $w_i$ is produced by computing the Euclidean distance between the randomly sampled embedding (denoted by $z_{s}$) and the best-matching embedding (denoted by $z_{q}$): $w_{i}=\| z_s - z_q \|_2^2$.
Then, the probabilistic contrastive loss $\mathcal{L}_{pcl}$ can be formulated by adjusting the pulling force of positive pairs with the normalized weighting parameters $\{w_i'\}_{i=1}^{L}, s.t. \sum_{i=1}^{N}w_i'=1$ as below:
\begin{equation}
\label{pcl}
\mathcal{L}_{pcl} = -\sum_{i=1}^L \log
\frac{w_i' \cdot e^{y_i \cdot z_i / \tau}} 
{w_i' \cdot e^{y_i \cdot z_i / \tau} + \frac{1}{L} \sum_{\substack{j=1 \\ j\neq i}}^L e^{ y_i \cdot z_j / \tau}}.
\end{equation}
Note, we balance the negative term with $1/L$, otherwise the negative term will be too large compared with the vanilla contrastive loss.

\renewcommand\arraystretch{1.25}
\begin{table*}[t]
\footnotesize
\caption{
Semantic image synthesis with \textbf{auto-regressive} models on ADE20K and CelebA-HQ, and text-to-image synthesis with \textbf{diffusion} models on CUB-200 and MS-COCO. 
[R] and [G] denote the results of reconstructed images, generated images with auto-regressive models or diffusion models.
}
\renewcommand\tabcolsep{2.25pt}
\centering 
\begin{tabular}{l||ccc||ccc||ccc||ccc} \hline
\multirow{2}{*}{\textbf{Models}}
& 
\multicolumn{3}{c||}{\textbf{ADE20K \cite{zhou2017ade20k} (Semantic)}} & 
\multicolumn{3}{c||}{\textbf{CelebA-HQ \cite{liu2015celebahq} (Semantic)}} &
\multicolumn{3}{c||}{\textbf{CUB-200 \cite{welinder2010caltech} (Text)}} & 
\multicolumn{3}{c}{\textbf{MS-COCO \cite{lin2014microsoft} (Text)}}
\\
\cline{2-13}
& FID[R]$\downarrow$ & PSNR[R]$\uparrow$ & FID[G]$\downarrow$ & FID[R]$\downarrow$ & PSNR[R]$\uparrow$ & FID[G]$\downarrow$ 
& FID[R]$\downarrow$ & PSNR[R]$\uparrow$ & FID[G]$\downarrow$ & FID[R]$\downarrow$ & PSNR[R]$\uparrow$ & FID[G]$\downarrow$
\\\hline 

\textbf{VQ-VAE} \cite{wang2018pix2pixhd} & 49.21 & \textbf{19.95} & 60.29         & 28.38  & \textbf{23.39} & 39.57 
& 20.89 & \textbf{21.07} & 26.32        & 32.48 & \textbf{19.12} & 38.84  \\

\textbf{VQ-GAN} \cite{esser2020taming} & 28.17 & 18.89 & 38.53        & 12.74  & 22.44 & 17.42                
& 13.49 & 20.88 & 17.43         & 18.58 & 18.86 & 23.75 
\\

\textbf{Gumbel-VQ} \cite{baevski2019vq}   & 26.42 & 18.08 & 37.51        & 12.03 & 20.92  & 16.78               
& 13.25 & 20.06 & 16.93        & 16.97 & 18.43 & 22.21    
\\

\rowcolor{mygray} \textbf{Reg-VQ}
& \textbf{23.69} & 18.44 & \textbf{34.47}
& \textbf{10.09} & 22.05 & \textbf{15.34}

& \textbf{10.84} & 20.39 & \textbf{14.14}
& \textbf{13.76} & 18.64 & \textbf{19.91}
  \\\hline
  
\end{tabular}
\label{tab_ar}
\end{table*}

\begin{figure*}[t]
\centering
\includegraphics[width=1.0\linewidth]{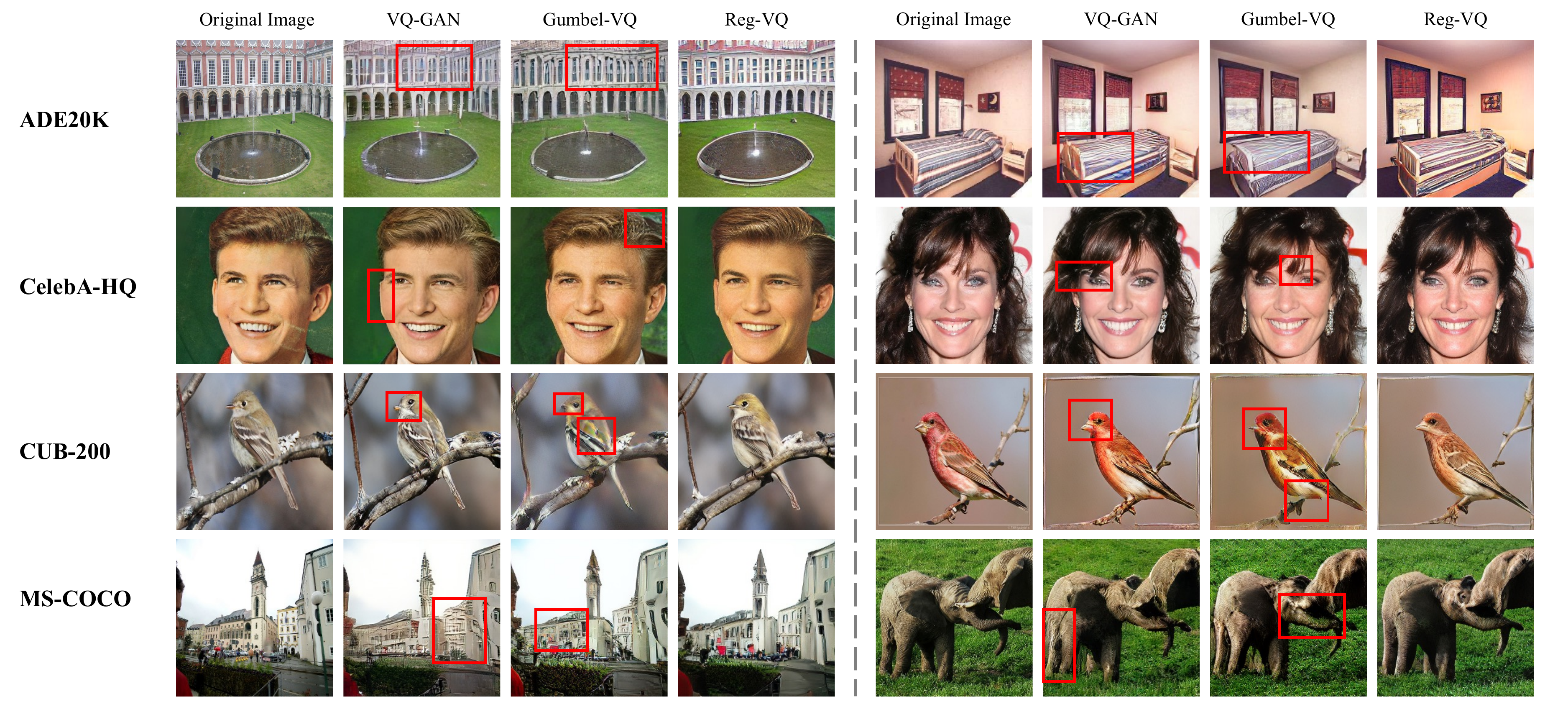}
\caption{
Reconstruction of images from four public datasets with different quantization methods: The red-color boxes highlight reconstruction artifacts.
}
\label{im_rec}
\end{figure*}

\section{Experiments}

\subsection{Experimental Settings}
\label{setting}

\textbf{Datasets}
We benchmark our method over multiple public datasets, including ADE20K \cite{zhou2017ade20k} and CelebA-HQ \cite{liu2015celebahq} for semantic image synthesis, CUB-200 \cite{welinder2010caltech} and MS-COCO \cite{lin2014microsoft} for text-to-image synthesis.

\textbf{Evaluation Metrics.}
We evaluate the vector quantization models by assessing their image reconstruction and image generation performance.
The image reconstruction performance is evaluated with several widely adopted evaluation metrics. 
Specifically, Fr{\'e}chet Inception Score (FID)~\cite{fid} is employed to evaluate the quality (perceptual similarity) of reconstructed images and generated images; Peak Signal-to-noise Ratio (PSNR) is employed to measure the accuracy (pixel-level similarity) of image reconstruction.

\textbf{Implementation Details.}
Following VQ-GAN \cite{esser2020taming}, the default feature size $H\times W$ and codebook size $N$ for all methods are set as $16\times 16$ and 1024, respectively~\footnote{Thus, the results reported in our paper are different from that in VQ-Diffusion whose default feature size and codebook size are $32\times 32$ and $8192$, respectively.}.
An image size of $256 \times 256$ is adopted for both image reconstruction and image generation.
A masking ratio of 40\% is adopted in our experiments by default.
More details of the experiments (e.g., hyper-parameters, network architectures) are provided in the supplemental material.

\begin{figure*}[htp!]
\centering
\includegraphics[width=1.0\linewidth]{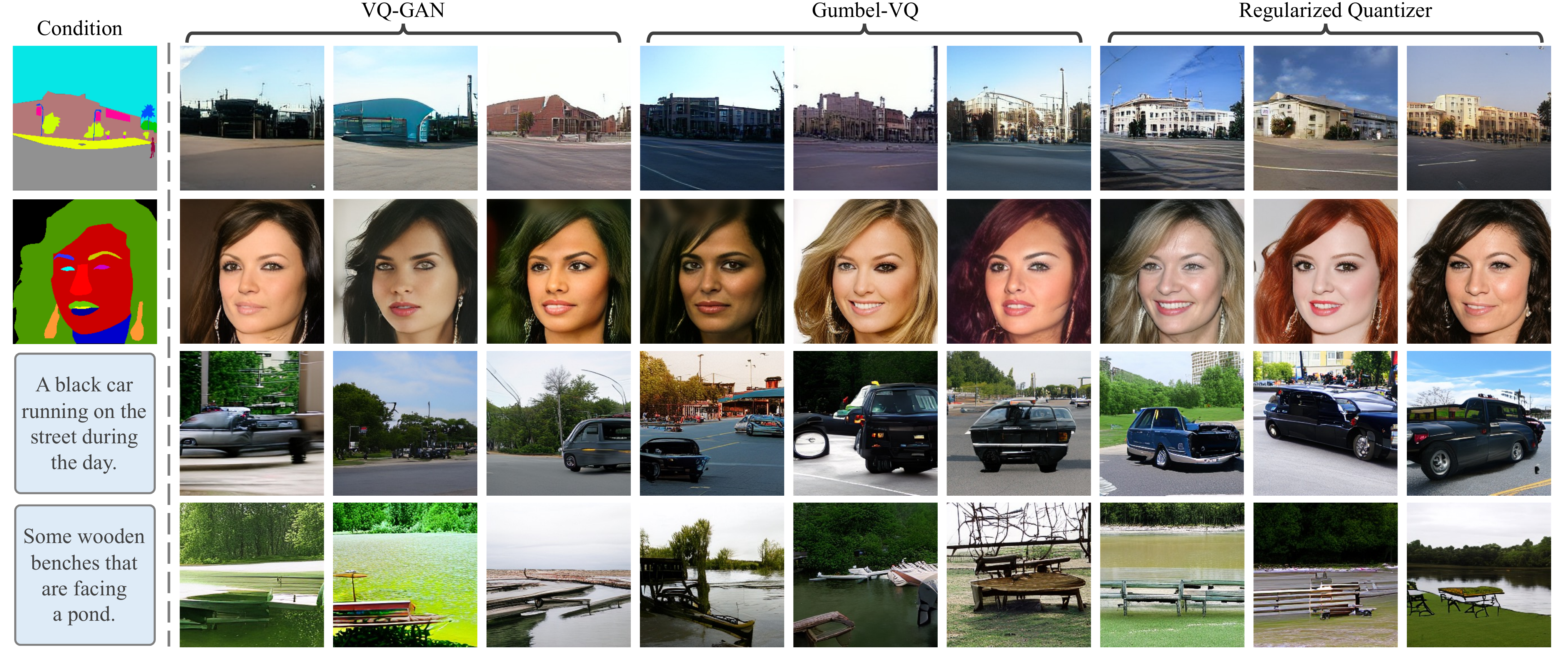}
\caption{
Semantic image synthesis and text-to-image synthesis. Three synthesis samples are shown for each condition under each quantization method.
}
\label{im_generation}
\end{figure*}

\subsection{Quantitative Evaluation}
Vector quantization performance can be evaluated by assessing their image reconstruction and image generation performance.
For image reconstruction, quantized image tokens are fed into the decoders of quantization models to recover the original images.
For image generation, auto-regressive model \cite{esser2020taming} \footnote{https://github.com/CompVis/taming-transformers} is employed for semantic image synthesis on ADE20K and CelebA-HQ; diffusion model \cite{gu2021vector} \footnote{https://github.com/cientgu/VQ-Diffusion} is employed for text-to-image synthesis on CUB-200 and MS-COCO.
Details of the used auto-regressive model and diffusion model are provided in the supplementary material.

Table~\ref{tab_ar} shows the image reconstruction \& generation results on ADE20K \& CelebA-HQ and CUB-200 \& MS-COCO.
VQ-GAN \cite{esser2020taming} is a deterministic quantization method, 
while Gumbel-VQ is a stochastic variant of VQ-GAN by employing Gumbel-Softmax for token sampling as in DALL-E \cite{ramesh2021zero}.
It can be observed that the proposed regularized quantization (Reg-VQ) outperforms all compared methods in terms of the reconstruction quality and generation quality as evaluated by FID[R] and FID[G], respectively.
VQ-GAN achieves relatively high reconstruction accuracy as evaluated by PSNR[R], as it naively selects the best-matching token for reconstruction.
However, the high reconstruction accuracy doesn't contribute to the image generation performance as reconstructing the original images is not the objective of image generation.
Instead, we can observe that it is the high \textbf{reconstruction quality} (as evaluated by FID[R]) that indicates a high generation performance (as evaluated by FID[G]), 
which can be further validated in the ensuing qualitative evaluation.

\subsection{Qualitative Evaluation}
We qualitatively compare the image reconstruction and generation performance of different methods as shown in Fig.~\ref{im_rec} and Fig.~\ref{im_generation}.
Regularized quantization (Reg-VQ) achieves the best reconstruction quality although its reconstructed images are not exactly aligned with original images in terms of detailed textures.
As suffering from codebook collapse or low codebook utilization,
both VQ-GAN and Gumbel-VQ present inferior reconstruction quality.
Regularized quantization also achieves superior synthesis quality on various image generation tasks (semantic image synthesis and text-to-image synthesis) and generative models (auto-regressive model and diffusion model) as illustrated in Fig. \ref{im_generation}.

\renewcommand\arraystretch{1.25}
\begin{table}[t!]
\small
\caption{
Ablation study on semantic image synthesis with auto-regressive models. VQ-GAN \cite{esser2020taming} serves as the baseline model.
`PriorReg', `MaskReg', `CL', `PCL' denote the prior distribution regularization, stochastic mask regularization, vanilla contrastive loss, and our probabilistic contrastive loss, respectively. The row in grey denotes the result of the standard regularized quantization.
}
\renewcommand\tabcolsep{3pt}
\centering 
\begin{tabular}{l||ccc} \hline
\multirow{2}{*}{\textbf{Models}}
& 
\multicolumn{3}{c}{\textbf{ADE20K \cite{zhou2017ade20k} (Semantic)}} 
\\
\cline{2-4}
& FID[R]$\downarrow$ & PSNR[R]$\uparrow$ & FID[G]$\downarrow$ 
\\\hline 

\textbf{Baseline \cite{esser2020taming}}  & 28.17 & 18.89 & 38.53       
\\

\textbf{+PriorReg}  & 25.92 & \textbf{18.98} & 36.57    
\\

\textbf{+PriorReg+MaskReg} & 25.11 & 18.56 & 35.03     
\\

\textbf{+PriorReg+MaskReg+CL}  & 24.21 & 18.49 & 34.91 
\\

\rowcolor{mygray}  \textbf{+PriorReg+MaskReg+PCL}  & \textbf{23.69} & 18.44 & \textbf{34.47}       
\\
  \hline
\end{tabular}

\label{tab_ablation}
\end{table}

\begin{figure*}[t]
\centering
\includegraphics[width=1.0\linewidth]{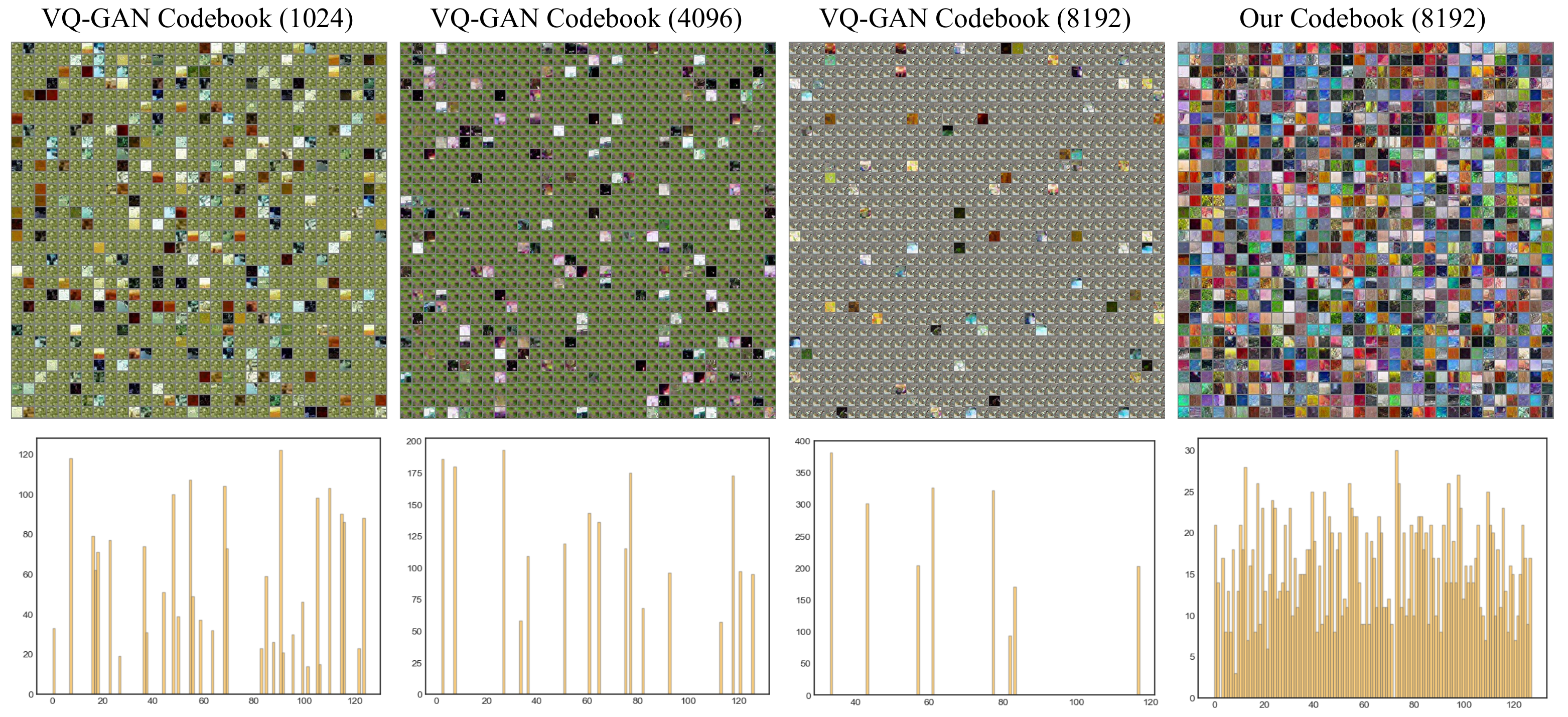}
\caption{
Visualization of codebook with different sizes (N=1024, 4096, 8192) on ADE20K. The first 1024 codebook embeddings are illustrated for all models.
Vanilla VQ-GAN (with deterministic quantization) suffers from severe codebook collapse with the increase of codebook size, while the regularized quantization achieves high codebook utilization consistently. The visualization of codebook on other datasets is provided in the supplementary material.
}
\label{im_codebook_size}
\end{figure*}

\begin{figure*}[t]
\centering
\includegraphics[width=1.0\linewidth]{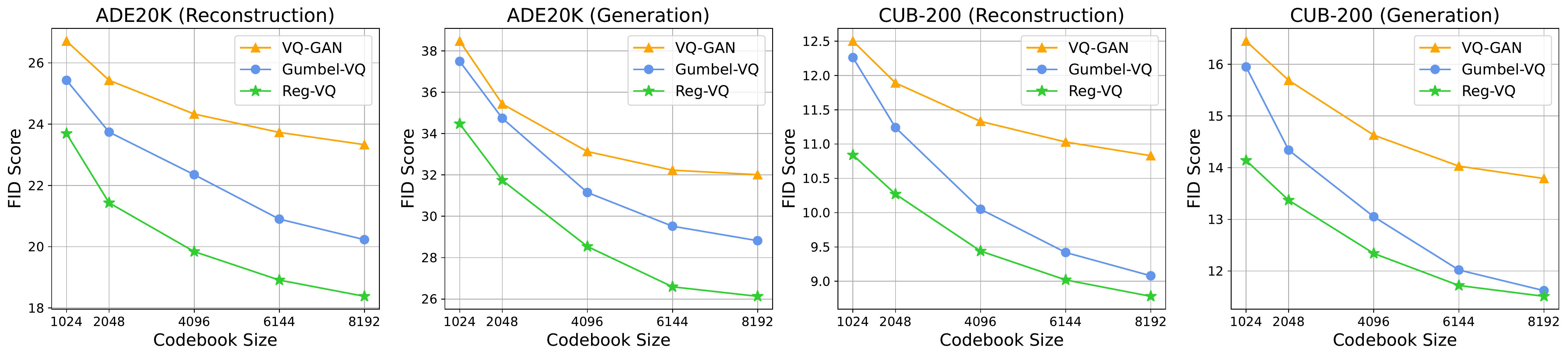}
\caption{
The effect of different codebook size on image reconstruction and image generation.
Auto-regressive model is employed for semantic image generation on ADE20K dataset and diffusion model is employed for text-to-image generation on CUB-200.
}
\label{im_codebook_graph}
\end{figure*}

\subsection{Ablation Study}
\label{ablation}

We conduct extensive ablation studies to evaluate the regularized quantization as shown in Table \ref{tab_ablation}.
The deterministic quantization method VQ-GAN \cite{esser2020taming} serves as the baseline model.
With the including of Prior distribution regularization (denoted by PriorReg),
the image reconstruction quality \& accuracy and generation quality are all improved substantially as evaluated by FID[R] \& PSNR[R] and FID[G].
The including of stochastic mask regularization (denoted by MaskReg) further improves the reconstruction and generation quality but degrades the reconstruction accuracy as shown in \textbf{+PriorReg+MaskReg}.
To mitigate the perturbed objective, contrastive loss (CL) is included and brings certain performance gains for image reconstruction and generation quality.
As a comparison, including the designed probabilistic contrastive loss (PCL) improves the reconstruction and generation performance by a larger margin.

We also study the effect of varying codebook size on ADE20K.
As shown in Fig. \ref{im_codebook_size}, the vanilla VQ-GAN with deterministic quantization suffers from severer codebook collapse with increasing of codebook size, while the proposed Regularized quantization still achieves high codebook utilization for large codebook size.
Fig. \ref{im_codebook_graph} quantitatively illustrates the reconstruction and generation performance with the different codebook sizes.
We can observe that the performance of regularized quantization (Reg-VQ) improves clearly with the increasing of codebook size, while the performance of VQ-GAN tends to be capped at a lower level.

\section{Conclusions}
\label{conclusion}
This paper presents a regularized quantization framework which achieves superior image quantization performance.
A prior distribution regularization is proposed to prevent the codebook collapse and low codebook utilization.
A stochastic mask regularization is designed to balance the inference stage misalignment and unperturbed reconstruction objective, and the masking ratio is analyzed comprehensively to yield the best performance.
To mitigate the perturbed reconstruction objective, a probabilistic contrastive loss is proposed to serve as a calibrated metric for elastic image reconstruction.
Quantitative and qualitative experiments show that regularized quantization enables to synthesize high-fidelity images for various generation tasks.

\section{Acknowledgments}

This project is funded by the Ministry of Education Singapore, under the Tier-2 project scheme with a project number MOE-T2EP20220-0003.
Fangneng Zhan and Christian Theobalt were supported by the ERC Consolidator Grant 4DReply (770784).

{\small
\bibliographystyle{ieee_fullname}
\bibliography{egbib}
}

\clearpage

\appendix

\section{Codebook Visualization}
\label{codebook}

The codebook is visualized by directly feeding the embeddings in the codebook into the decoder $G$ to generate codebook image patches.
Note the decoder $G$ is a fully convolution network and thus the input features to the decoder can have any sizes.
For all models, the first 1024 codebook embeddings (i.e., indices 0-1023) are visualized to form a image with $32\times 32$ patches.

We visualize the codebook of VQ-GAN and our regularized quantizer on CelebA-HQ \cite{liu2015celebahq}, CUB-200 \cite{welinder2010caltech}, and MS-COCO \cite{lin2014microsoft} as shown in Fig. \ref{im_codebook}.



\section{Experiment Details}
\label{experiment}

Both vector quantization methods and generative models are optimized via AdamW~\cite{loshchilov2017decoupled} solver ($\beta_{1}=0.9$ and $\beta_{2}=0.95$) with a learning rate of 1.5$e$-4. All experiments are conducted on 4 Tesla V100 GPUs with a batch size of 40.

For vector quantization, the feature size $H\times W$ is set as $16\times 16$ by default; the default codebook size $N$ and embedding dimension are set as 1024 and 256. 
The encoder and decoder in regularized quantizer follow the default structure of VQ-GAN \cite{esser2020taming}.
The temperature parameter in Gumbel-Softmax is set as 0.9 by default.
The training epochs for vector quantization on ADE20K, CelebA-HQ, CUB-200, and MS-COCO are 100, 60, 300, 50, respectively, for all models.

For auto-regressive modeling, the Transformer used in Esser \emph{et al.} \cite{esser2020taming,zhan2022auto} \footnote{https://github.com/CompVis/taming-transformers} is selected as the code base with the default setting.
Specifically, the vocabulary size, embedding number and input sequence length are 1024, 1024 and 512, respectively; the numbers of transformer blocks and attention head are 24 and 16, respectively.
For the task of semantic image synthesis, the \textit{semantic ids} of semantic maps are directly used as the conditional tokens for the Transformer.
The training epoch for the auto-regressive model is 50 for all datasets.

For the diffusion modeling, Denoising Diffusion Probabilistic
Model (DDPM) in VQ-Diffusion \cite{gu2021vector} \footnote{https://github.com/cientgu/VQ-Diffusion} is selected as the code base. Specifically, VQ-Diffusion-B (Base) which has 19 transformer blocks with dimension of 1024 is employed to estimate the token distribution.
The training epoch for the diffusion model is 100 for all datasets.

\section{Limitations and Future Work}
\label{limitation}

Current quantization models train the encoder and decoder with the same learning objective.
However, for tokenized image synthesis, the encoder and decoder in quantization models actually have different objectives: the encoder aims to learn accurate discrete representation, while the decoder aims to generate realistic images.
Thus, training them with the same objective tends to be sub-optimal and will constrain the quantization and generation performance.
In the future, we will explore to design separated learning objective for the encoder and decoder for optimal quantization and generation performance.
Besides, we can also explore the performance with different prior distribution, e.g., Gaussian distribution.

\section{Ethical Considerations} 
\label{ethical}

The proposed quantization method aims to synthesize realistic images.
There would be negative impacts if it is combined with generation methods for certain illegal purpose such as image forgery.

\section{More Qualitative Results}
\label{qualitative}
\begin{figure*}[t]
\centering
\includegraphics[width=1.0\linewidth]{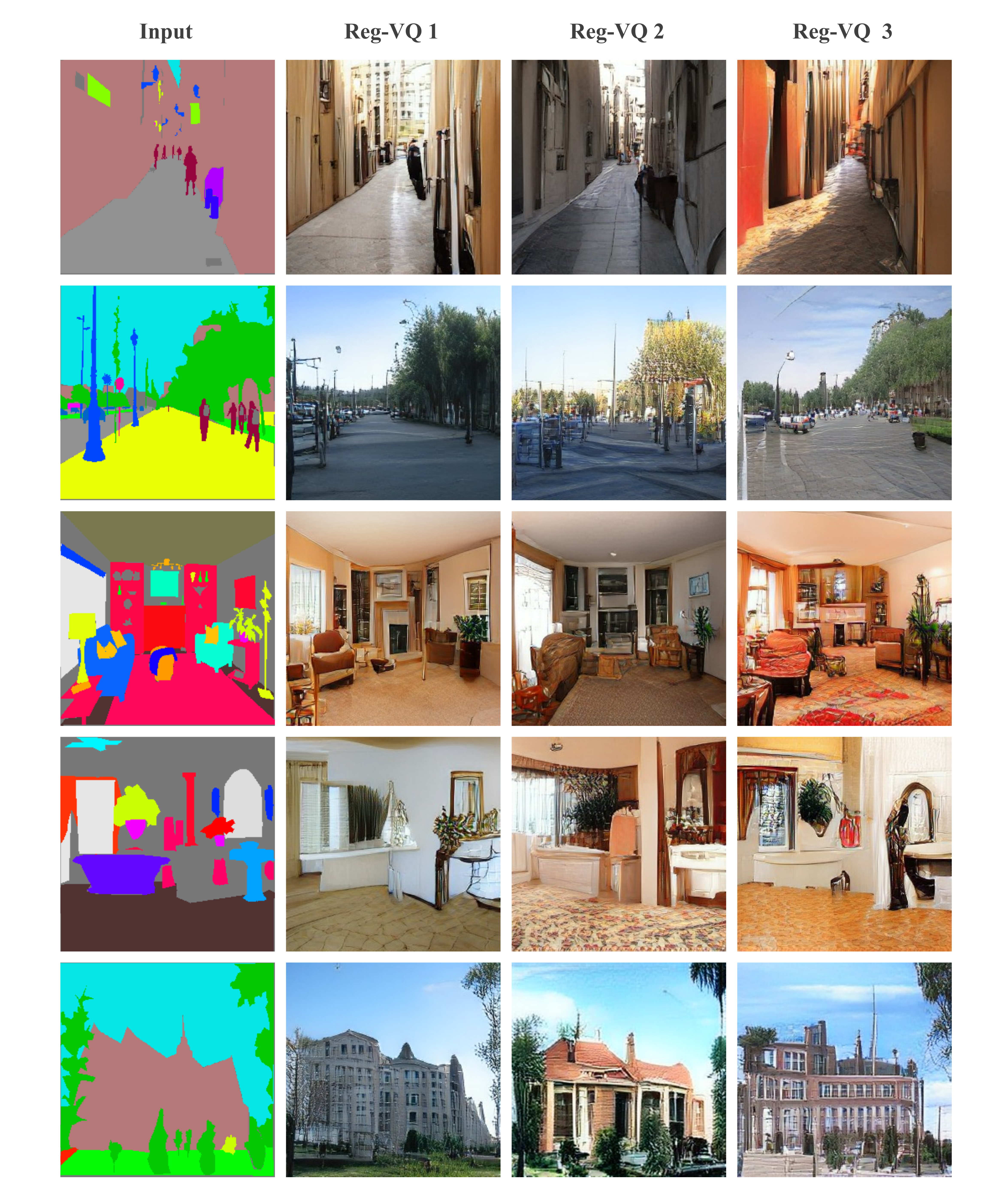}
\caption{
Semantic image synthesis on ADE20K with regularized quantizer and auto-regressive model.
}
\label{im_ade20k}
\end{figure*}

\begin{figure*}[t]
\centering
\includegraphics[width=1.0\linewidth]{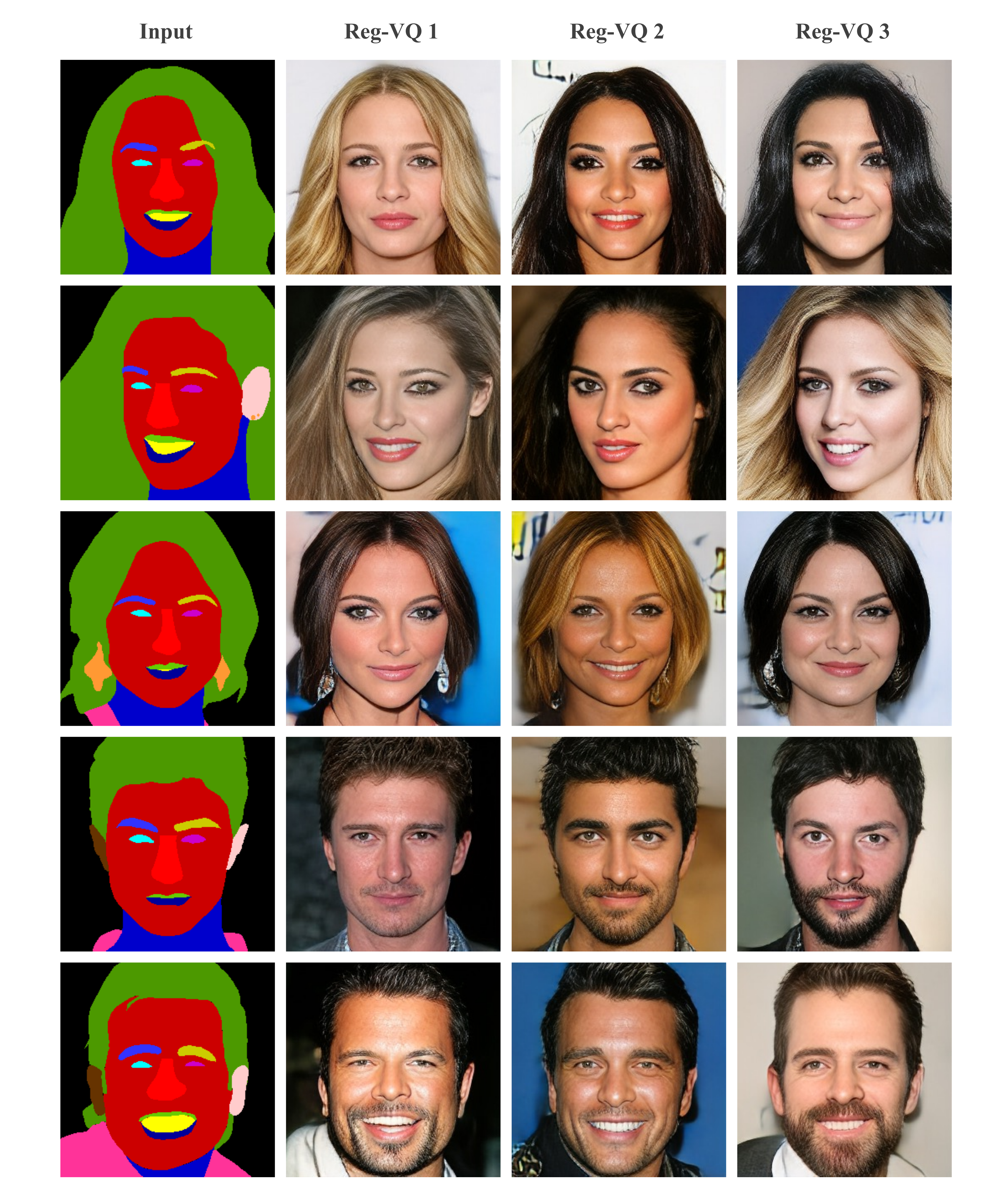}
\caption{
Semantic image synthesis on CelebA-HQ with regularized quantizer and auto-regressive model.
}
\label{im_celebahq}
\end{figure*}

\begin{figure*}[t]
\centering
\includegraphics[width=1.0\linewidth]{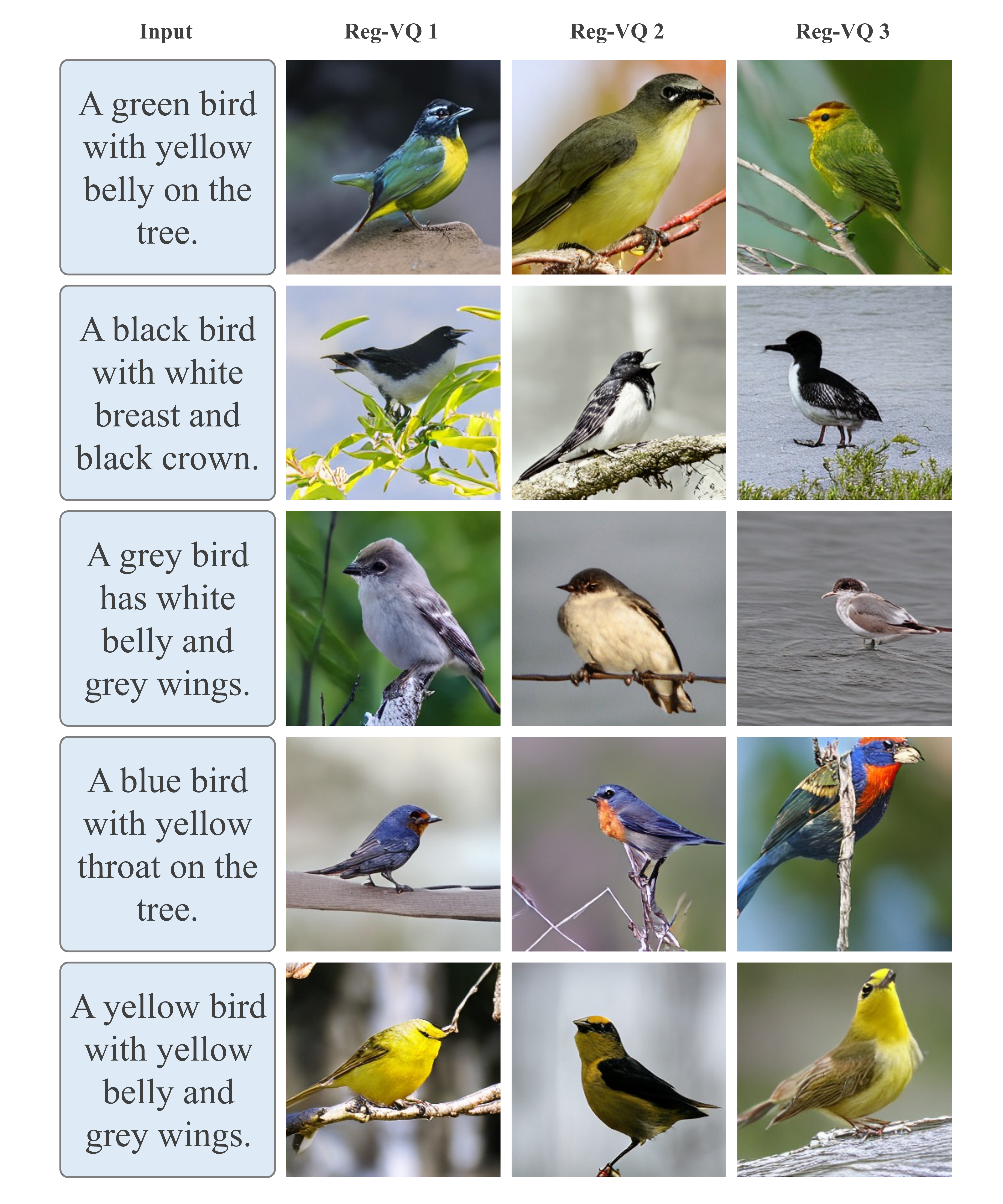}
\caption{
Text-to-image synthesis on CUB-200 with regularized quantizer and diffusion model.
}
\label{im_cub200}
\end{figure*}

\begin{figure*}[t]
\centering
\includegraphics[width=1.0\linewidth]{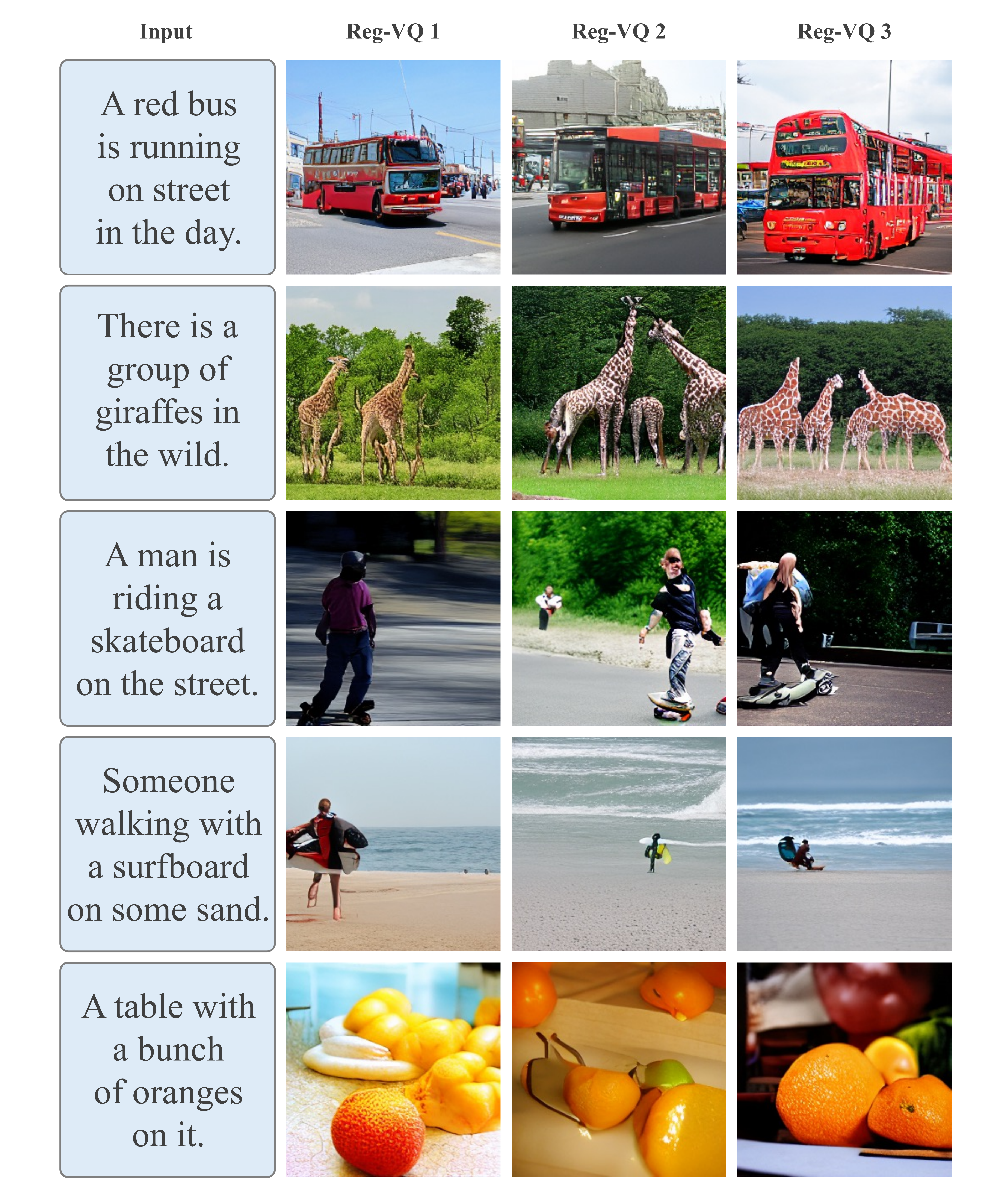}
\caption{
Text-to-image synthesis on MS-COCO with regularized quantizer and diffusion model.
}
\label{im_mscoco}
\end{figure*}

We provide more image translation results including Figs. \ref{im_ade20k}, \ref{im_celebahq} for semantic image synthesis, and Figs. \ref{im_cub200}, \ref{im_mscoco} for text-to-image synthesis.

\begin{figure*}[t]
\centering
\includegraphics[width=1.0\linewidth]{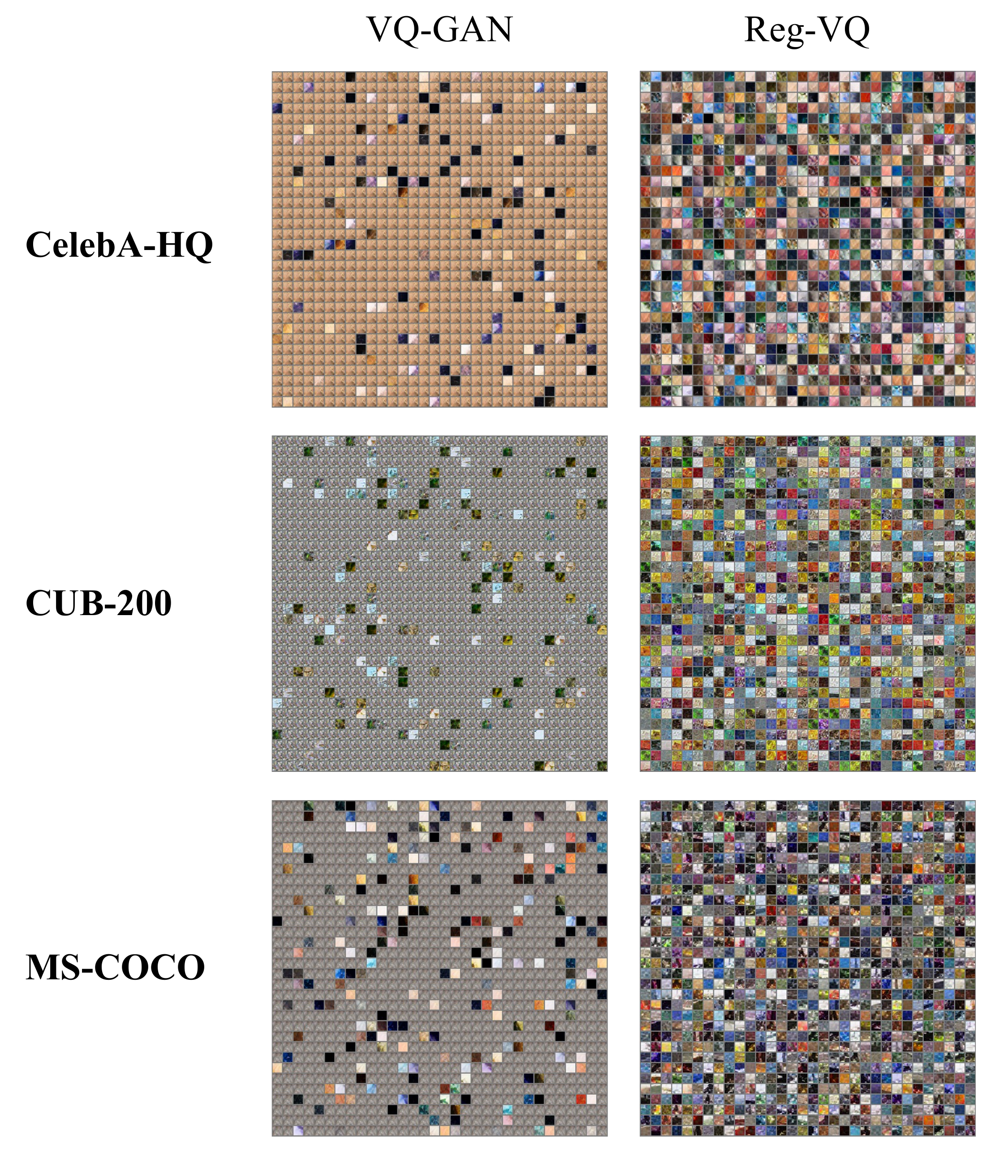}
\caption{
The visualization of codebook of VQ-GAN and regularized quantizer on CelebA-HQ, CUB-200, and MS-COCO.
}
\label{im_codebook}
\end{figure*}


\end{document}